\documentclass{article}

\PassOptionsToPackage{numbers, compress}{natbib}
\usepackage[preprint]{neurips_2026}

\usepackage[utf8]{inputenc}
\usepackage[T1]{fontenc}
\usepackage[hidelinks]{hyperref}
\usepackage{url}
\usepackage{booktabs}
\usepackage{amsfonts,amsmath,amssymb,amsthm}
\usepackage{nicefrac}
\usepackage{microtype}
\usepackage{graphicx}
\usepackage{multirow}
\usepackage{xcolor}
\usepackage{placeins}

\newtheorem{proposition}{Proposition}

\title{When Does an $E(2)$-Steerable CNN Guarantee Invariant Galaxy Classification?}

\author{%
  Alejandro Treny Ortega\\
  University College London\\
  \texttt{alejandro.ortega.26@ucl.ac.uk}
}

\begin{document}

\maketitle

\begin{abstract}
Rotating a radio-telescope cutout does not change the galaxy in it, so an $E(2)$-steerable classifier is expected to keep its label. In Scaife and Porter's released MiraBest model, roughly one galaxy in ten changes class anyway. Its dense head flattens a spatial grid, so it can weight positions that a rotation swaps differently; padding $150^2$ cutouts by one pixel also shifts the pre-decimation grid from odd to even before stride-two pooling, raising the relative error at the first pooling layer from $1.07\times 10^{-7}$ to $24.9\%$. On MiraBest ($N=833$, five-fold CV), flattened models flip on $9.8$--$12.5\%$ of galaxies under the eight quarter-turns and reflections ($D_4$; worst-case accuracy $84.4\%$), while global average pooling removes every flip but costs $5.2$--$10.2$~pp. In between, compact exact-$D_4$ readouts with $113.7\times$ fewer head weights reach $89.7$--$91.8\%$ in one pass with zero exact flips under longer training. Their weak point is position: after a $12$-pixel zero-padded cardinal shift, fixed-centre masks lose $12.7$--$27.5$~pp. Letting the same smooth Gaussian rings follow the activation centroid keeps exact $D_4$ invariance and holds accuracy at $89.7$--$90.1\%$ across $0$--$12$-pixel shifts, $11.5\pm 2.2$~pp above fixed-centre rings at $12$ pixels.
\end{abstract}

\section{Introduction}
\label{sec:intro}

Turn a radio-telescope cutout of a galaxy by $90^\circ$. Every pixel moves, yet the physical source on the sky is unchanged. Fanaroff and Riley~\cite{fanaroff1974morphology} separated edge-darkened Class~I (FRI) from edge-brightened Class~II (FRII) radio sources by where their brightest extended emission sits relative to the source extent (Fig.~\ref{fig:mechanism}a). In survey catalogues from LOFAR, EMU, or SKA, a label that flips under a quarter-turn or reflection injects orientation-dependent noise into morphology counts.

Group-equivariant~\cite{cohen2016group} and $E(2)$-steerable CNNs~\cite{weiler2019general,cesa2022program,bronstein2021geometric} were adopted in radio astronomy to build that symmetry into the network itself~\cite{scaife2021fanaroff,bowles2021e2,pandya2023e2,porter2023mirabest}, alongside other MiraBest baselines~\cite{bowles2021attention,mohan2022quantifying,slijepcevic2024radio,drozdova2025radio}. Scaife and Porter~\cite{scaife2021fanaroff} built $E(2)$-steerable LeNets~\cite{lecun1998gradient} on MiraBest, listing \emph{Global Average Pool} in Table~4 and stating guaranteed orientation-invariant classification in \S5.2 (while noting masking, interpolation, and discretisation limits in \S3 and \S6.4).

Reading the pinned public code (commit~\texttt{90d104b3}~\cite{scaife2021code}), two details stand out. The dense classifier does not receive a globally pooled vector; it receives a flattened $36\times 36$ feature map. Before the first convolution, every $150^2$ cutout is also padded asymmetrically to $151^2$. The convolutional trunk is equivariant by construction, yet its sampling grid breaks exact equivariance and its readout discards the guarantee. Two questions follow: where exactly does $D_4$ symmetry break, and how much spatial layout does a classifier need to stay accurate without giving up invariance~\cite{zhang2019making,xu2021group,edixhoven2023using,rahman2025group}? We test both failure modes in code and theory, compare compact invariant readouts with full-group logit averaging~\cite{puny2022frame}, and then move the galaxies off-centre. Code and per-galaxy CSVs are available at \url{https://github.com/aleetreny/steerable-galaxy-invariance}.

\section{Two Places Where Exact $D_4$ Invariance Breaks}
\label{sec:theory}

Tracing \texttt{models.py} and \texttt{main.py} in commit~\texttt{90d104b3}~\cite{scaife2021code} (App.~\ref{app:proofs}) gives the forward path of \texttt{DNSteerableLeNet}: $x \to \text{Pad}_{151} \to \text{Conv}_1 \to \text{ReLU} \to \text{PoolAA}_1 \to \text{Conv}_2 \to \text{ReLU} \to \text{PoolAA}_2 \to \text{GPool} \to \text{flat} \to \text{MLP}$. Let $D_4$ be the eight exact symmetries of the square pixel lattice. After steerable convolutions, \texttt{GroupPooling} takes the maximum over orientation fibres at each pixel, leaving a non-negative feature map $F(x)\in\mathbb{R}^{C\times S\times S}$ ($16\times 36^2 = 20{,}736$ entries). On a $D_4$-equivariant trunk, each $g\in D_4$ still permutes those pixels via $P_g \in \{0,1\}^{S^2\times S^2}$, so $\operatorname{vec}(F_c(g\cdot x)) = P_g \operatorname{vec}(F_c(x))$.

\begin{proposition}[Exact $D_4$-Invariant Affine Readout]
\label{prop:readout}
Let $h(F)=W\operatorname{vec}(F)+b$ be an affine layer with channel-$c$ weight slice $w_{j,c}\in\mathbb{R}^{S^2}$ into unit $j$, and define $(g\cdot F)_c = \operatorname{vec}^{-1}(P_g\operatorname{vec}(F_c))$. Then $h(g\cdot F)=h(F)$ for all $F\in\mathbb{R}^{C\times S\times S}$ and $g\in D_4$ if and only if $P_g^\top w_{j,c} = w_{j,c}$ for all $j,c,g$ ($w_{j,c}$ is constant on each $D_4$-orbit). Thus, for a $D_4$-equivariant trunk, any fixed pooling matrix $M\in\mathbb{R}^{K\times S^2}$ with $MP_g=M$ is sufficient for exact $D_4$ invariance in real arithmetic for any evaluation-mode MLP.
\end{proposition}

\emph{Proof sketch.} Applying $g$ shifts unit $j$'s pre-activation by $\sum_c ((P_g^\top - I)w_{j,c})^\top\operatorname{vec}(F_c)$, which vanishes on all of $\mathbb{R}^{C\times S\times S}$ if and only if $(P_g^\top - I)w_{j,c}=0$ (full proof in App.~\ref{app:proofs}). \hfill$\square$

\begin{figure}[!t]
  \centering
  \includegraphics[width=0.73\textwidth]{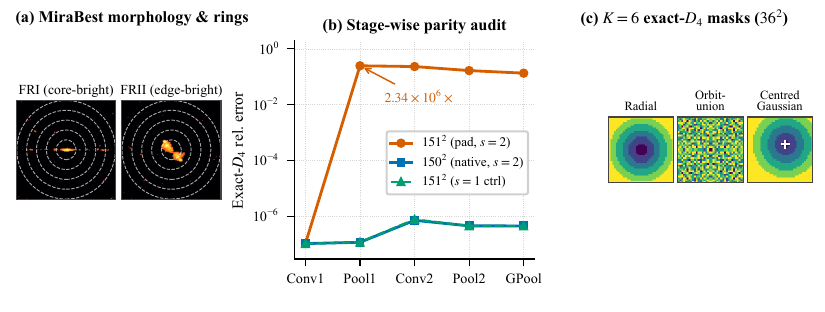}
  \caption{Morphology, parity audit, and exact-$D_4$ readouts. (a)~MiraBest FRI and FRII cutouts with $K=6$ schematic rings. (b)~Stage-wise exact-$D_4$ relative error (shared $D_4$ weights): padding $150\to 151$ jumps $2.34\times 10^6\times$ at \texttt{Pool1}, whereas native $150^2$ and stride-1 ($N=4$) stay at $10^{-7}$--$10^{-6}$. (c)~$K=6$ masks on $36^2$ (Centred Gaussian shows the argmax ring around~$+$).}
  \label{fig:mechanism}
\end{figure}

\textbf{Designing compact invariant readouts.} Flattening into $120$ hidden units allocates $2{,}498{,}774$ weights to the head against $8{,}514$ convolutional weights in $D_4$, leaving it free to weight rotated positions differently. Global average pooling (GAP, $K=1$, $12{,}374$ weights) satisfies $MP_g=M$ trivially, but a global mean loses where the emission sits. We asked how much of that layout the classifier needs. We match five exact-$D_4$ readouts at $96$ features ($16\times 6$) and $21{,}974$ head weights, $113.7\times$ fewer than flattening (Fig.~\ref{fig:mechanism}c):
(i)~Global statistics pools six orderless summaries per channel (mean, std, max, $q_{75}$, $q_{90}$, RMS);
(ii)~Orbit-union groups exact $D_4$ orbits on $36^2$ into six disjoint masks without radial ordering (mean radii $12.7$--$13.9$~px);
(iii)~Radial averages over six disjoint concentric annuli centred at $c_0=17.5$;
(iv)~Fixed Gaussian replaces those hard edges with six smooth overlapping Gaussian rings at fixed $c_0=17.5$;
(v)~Centred Gaussian uses the same ring centres and width $\sigma$, anchored to the $D_4$-equivariant activation centre of mass $c(F)=\sum_u w(u)\,u$ of $\bar{F}(u)=\frac{1}{C}\sum_c \operatorname{ReLU}(F_c(u))+10^{-6}$, keeping $\|g\cdot u - c(g\cdot F)\|_2 = \|u - c(F)\|_2$ exact in real arithmetic (App.~\ref{app:proofs}).

\textbf{How one pixel breaks stride-2 parity.} In \texttt{main.py}, native $150^2$ cutouts are padded asymmetrically to $151^2$ with \texttt{Pad(0,0,1,1)}. The pool applies a $2\times 2$ max filter, then a $5\times 5$ Gaussian blur sampled at stride two~\cite{zhang2019making}; source inspection confirms the same two-stage operator in \texttt{escnn}~\cite{cesa2022program} (App.~\ref{app:proofs}). On a native $150^2$ image, the $5\times 5$ convolution (padding~1) and $2\times 2$ max pool leave a $147^2$ map. Because $147$ is \emph{odd}, reflecting $i\mapsto 146-i$ maps even indices onto even indices, preserving the stride-2 grid ($74\to 72\to 71\to 36$). Adding one pixel turns $147$ into $148$ (\emph{even}): reflection $i\mapsto 147-i$ now swaps even and odd indices inside the trunk~\cite{xu2021group,edixhoven2023using}. Equivariant downsampling~\cite{xu2021group,rahman2025group} or polyphase sampling~\cite{chaman2021truly} avoids this phase mismatch.

\section{Experimental Setup and Controls}
\label{sec:protocol}

We evaluate on MiraBest~\cite{porter2023mirabest}, with $833$ \emph{Confident} ($397$ FRI, $436$ FRII) and $389$ \emph{Uncertain} ($194$/$195$) cutouts (checksums, normalisation, and optimiser details in App.~\ref{app:shifts}). To isolate parity before training, we pass $N=16$ cutouts through $D_4$ and $D_{16}$ trunks with shared untrained weights on $150^2$ and post-padding $151^2$, plus an equal-batch ($N=4$) stride-1 control.

For classification, we run stratified five-fold cross-validation over the $833$ Confident galaxies on $150^2$ under a short budget ($60$ epochs, $300$ steps, three seeds for Global statistics, Orbit-union, and Radial) and a longer budget ($120$ epochs, batch size $64$, $1{,}080$ steps) across all six $D_4$ heads, alongside a Vanilla CNN ($3\times 3$ BlurPool~\cite{zhang2019making}) and exact 8-view $D_4$-TTA of Flatten~\cite{puny2022frame}. To tell trunk features from head capacity, we refit five heads on frozen Flatten and Radial trunks under the short schedule and a longer $3.4\text{k}$-step 8-view protocol. Survey cutouts are rarely perfectly centred, so we also apply zero-padded even cardinal shifts $\Delta\in\{0,2,4,8,12\}$ pixels before normalisation ($3{,}332$ evaluations per nonzero $\Delta$, with exact $D_4$ flips checked on each) and evaluate short-budget models on Uncertain. We report accuracy, unclipped NLL, 10-bin ECE, exact $D_4$ and 32-view PIL ($22.5^\circ\times$ reflection) flips, WC-$D_4$, Holm-adjusted exact McNemar $p$-values, and conditional paired bootstrap $95\%$ CIs.

\section{Results}
\label{sec:results}

\begin{table}[!t]
\centering
\caption{Five-fold cross-validation on MiraBest Confident ($N=833$, native $150^2$). Top: $120$-epoch training ($1{,}080$ steps) and exact 8-view $D_4$-TTA. Bottom: $60$-epoch training ($300$ steps) and three-seed means ($2{,}499$ evaluations per head). Acc: accuracy (galaxies correct); flip columns: galaxies whose label changes in at least one view (\% of $833$); WC-$D_4$: correct in all eight exact $D_4$ views; `---' indicates not evaluated. Bold: best single-pass exact-$D_4$ head in each column of the 120-epoch block.}
\label{tab:main_cv}
\scriptsize
\setlength{\tabcolsep}{1.7pt}
\begin{tabular}{@{}lrrcccccc@{}}
\toprule
\textbf{Model ($150^2$)} & \textbf{Head Params} & \textbf{Passes} & \textbf{Acc (\%)} & \textbf{WC-$D_4$} & \textbf{NLL} & \textbf{ECE} & \textbf{Exact $D_4$} & \textbf{PIL32} \\
\midrule
\multicolumn{9}{@{}l}{\textit{120 epochs, 1,080 steps (mean best validation epochs in App.~\ref{app:stats})}} \\
$D_4$ Flatten & $2{,}498{,}774$ & $1\times$ & $92.20\ (768)$ & $84.39$ & $0.241$ & $0.038$ & $104\ (12.5\%)$ & $158\ (19.0\%)$ \\
$D_4$ Flatten + $D_4$-TTA & $2{,}498{,}774$ & $8\times$ & $92.80\ (773)$ & $92.80$ & $0.196$ & $0.023$ & $0\ (0.0\%)$ & --- \\
$D_4$ GAP & $12{,}374$ & $1\times$ & $87.03\ (725)$ & $87.03$ & $0.320$ & $0.027$ & $0\ (0.0\%)$ & $64\ (7.7\%)$ \\
$D_4$ Orbit-union ($K=6$) & $21{,}974$ & $1\times$ & $\mathbf{91.84}\ (765)$ & $\mathbf{91.84}$ & $\mathbf{0.218}$ & $0.025$ & $0\ (0.0\%)$ & $67\ (8.0\%)$ \\
$D_4$ Radial ($K=6$) & $21{,}974$ & $1\times$ & $90.28\ (752)$ & $90.28$ & $0.279$ & $0.029$ & $0\ (0.0\%)$ & $45\ (5.4\%)$ \\
$D_4$ Fixed Gaussian ($K=6$) & $21{,}974$ & $1\times$ & $91.00\ (758)$ & $91.00$ & $0.250$ & $0.024$ & $0\ (0.0\%)$ & $\mathbf{39}\ (4.7\%)$ \\
$D_4$ Centred Gaussian ($K=6$) & $21{,}974$ & $1\times$ & $89.68\ (747)$ & $89.68$ & $0.265$ & $\mathbf{0.020}$ & $0\ (0.0\%)$ & $55\ (6.6\%)$ \\
\midrule
\multicolumn{9}{@{}l}{\textit{60 epochs, 300 steps and three-seed check}} \\
Vanilla CNN (BlurPool) & $2{,}498{,}774$ & $1\times$ & $91.84\ (765)$ & $83.55$ & $0.239$ & $0.025$ & $111\ (13.3\%)$ & $165\ (19.8\%)$ \\
$D_4$ Flatten & $2{,}498{,}774$ & $1\times$ & $90.88\ (757)$ & $85.59$ & $0.226$ & $0.026$ & $82\ (9.8\%)$ & $131\ (15.7\%)$ \\
$D_4$ GAP & $12{,}374$ & $1\times$ & $80.67\ (672)$ & $80.67$ & $0.446$ & $0.074$ & $0\ (0.0\%)$ & $99\ (11.9\%)$ \\
\multicolumn{9}{@{}l}{\textit{Three-seed means:} Global statistics $85.23{\pm}0.52\%$ (PIL $14.0\%$); Orbit-union $86.99{\pm}0.80\%$ ($8.8\%$); Radial $87.92{\pm}0.49\%$ ($5.6\%$).} \\
\bottomrule
\end{tabular}
\end{table}

\textbf{Where symmetry breaks in the trunk (Fig.~\ref{fig:mechanism}b, App.~Table~\ref{tab:app_audit}).} The audit pins the break to a single layer. On native $150^2$ cutouts, exact-$D_4$ error stays at float32 noise from \texttt{Conv1} ($1.07\times 10^{-7}$) through \texttt{GPool} ($4.58\times 10^{-7}$). On post-padding $151^2$ with the same weights, \texttt{Conv1} stays exact, then \texttt{Pool1} jumps $2.34\times 10^6\times$ to $24.93\%$. Stride $1$ on the same padded grid ($N=4$) restores float32 noise ($1.20\times 10^{-7}$) in both groups. Before any training, every invariant head on $150^2$ keeps logit variation below $1.05\times 10^{-6}$, whereas flattening already flips $12.5\%$ ($D_4$) to $62.5\%$ ($D_{16}$).

\begin{figure}[!t]
  \centering
  \includegraphics[width=0.71\textwidth]{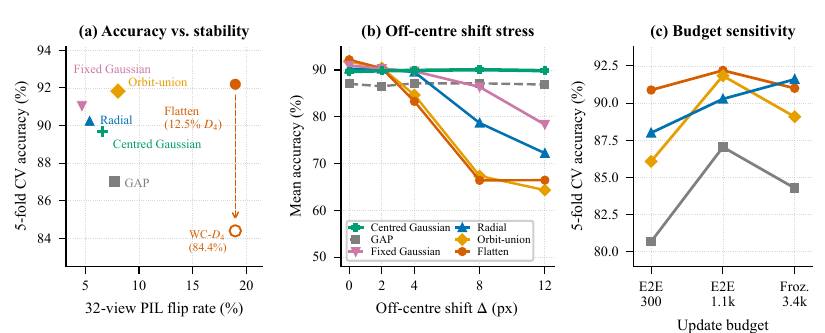}
  \caption{Classification accuracy under changes in orientation and cutout position ($N=833$). (a)~Long-budget CV accuracy vs.\ 32-view PIL flip rate (arrow: Flatten canonical to WC-$D_4$). (b)~Mean accuracy over four cardinal shifts ($\Delta=0$--$12$~px): Fixed Gaussian recovers $+6.1$~pp over hard Radial at $12$~px, and Centred Gaussian adds $+11.5$~pp more, staying near $90\%$. (c)~Training regimes (short E2E, long E2E, 8-view frozen Flatten trunk).}
  \label{fig:pareto}
\end{figure}

\textbf{Accuracy, orientation stability, and training budget (Table~\ref{tab:main_cv}, Fig.~\ref{fig:pareto}a,c).} Under $300$ steps, flattening looks attractive: $D_4$ Flatten reaches $90.88\%$ accuracy, yet flips class on $82/833$ galaxies ($9.8\%$), nearly a third of them at over $80\%$ confidence, so only $85.59\%$ are correct in all eight views ($83.55\%$ for Vanilla). The obvious fix, GAP, removes every exact flip but costs $10.20$~pp ($p=1.5\times 10^{-10}$). Keeping some layout helps: Radial ($87.92\%$) beats Global statistics ($85.23\%$) on all three seeds and cuts PIL32 flips from $14.0\%$ to $5.6\%$.

Training for $120$ epochs ($1{,}080$ steps) narrows the gap, though not in the same way for every model. Flatten reaches its minimum validation loss early (mean epoch $36.2$) and still flips $104$ galaxies ($12.5\%$, WC-$D_4=84.39\%$); 8-view $D_4$-TTA removes those flips ($92.80\%$) at the price of eight passes. Compact invariant heads keep improving until epochs $59$--$96$. On centred cutouts, Orbit-union comes within $0.36$~pp of Flatten (conditional $95\%$ CI $[-2.52, +1.80]$, $p=0.824$) with zero exact flips, Fixed Gaussian reaches $91.00\%$ with the lowest observed PIL32 flip rate ($4.7\%$), and Centred Gaussian reaches $89.68\%$ with the lowest ECE ($0.020$). The frozen-trunk refits agree: on Flatten's own trunk, Radial rises from $88.00\%$ to $91.60\%$ under the longer 8-view protocol (Table~\ref{tab:diag}).

\begin{table}[!t]
\centering
\caption{Off-centre shift accuracy (\%) at $\Delta{=}0,4,8,12$~px (invariant heads: $0.0\%$ $D_4$ flips; Flatten: $12.5{\to}33.2\%$), frozen Flatten-trunk refits ($N{=}833$), and short-budget Uncertain ($N{=}389$, $1{,}945$ evaluations per head; accuracy and flip rates in \%), where invariant heads still never flip. Bold: best exact-$D_4$ head per column (ties unmarked); `---' indicates not evaluated.}
\label{tab:diag}
\scriptsize
\setlength{\tabcolsep}{2.5pt}
\begin{tabular}{@{}lccccccccc@{}}
\toprule
& \multicolumn{4}{c}{\textbf{Off-Centre Shift Acc (\%)}} & \multicolumn{2}{c}{\textbf{Frozen Trunk (Acc / NLL)}} & \multicolumn{3}{c}{\textbf{Uncertain (Short, $N{=}389$)}} \\
\cmidrule(lr){2-5}\cmidrule(lr){6-7}\cmidrule(l){8-10}
\textbf{Head} & $\Delta{=}0$ & $\Delta{=}4$ & $\Delta{=}8$ & $\Delta{=}12$ & $300$ steps & $3.4\text{k}$ steps & Acc & $D_4$ & PIL32 \\
\midrule
$D_4$ Flatten & $92.20$ & $83.28$ & $66.39$ & $66.45$ & $91.36$ / $0.217$ & $91.00$ / $0.351$ & $64.5$ & $21.9$ & $33.7$ \\
$D_4$ GAP & $87.03$ & $87.15$ & $87.15$ & $86.88$ & $83.91$ / $0.474$ & $84.27$ / $0.366$ & $63.4$ & $0.0$ & $16.2$ \\
$D_4$ Orbit-union & $\mathbf{91.84}$ & $84.66$ & $67.35$ & $64.32$ & $87.27$ / $0.315$ & $89.08$ / $0.296$ & $63.4$ & $0.0$ & $16.5$ \\
$D_4$ Radial & $90.28$ & $89.56$ & $78.66$ & $72.21$ & $\mathbf{88.00}$ / $\mathbf{0.286}$ & $\mathbf{91.60}$ / $\mathbf{0.251}$ & $63.3$ & $0.0$ & $\mathbf{11.0}$ \\
$D_4$ Fixed Gaussian & $91.00$ & $89.74$ & $86.37$ & $78.33$ & --- & --- & --- & --- & --- \\
$D_4$ Centred Gaussian & $89.68$ & $\mathbf{89.92}$ & $\mathbf{90.07}$ & $\mathbf{89.86}$ & --- & --- & --- & --- & --- \\
\bottomrule
\end{tabular}
\end{table}

\textbf{What happens when the galaxy is off-centre (Table~\ref{tab:diag}, Fig.~\ref{fig:pareto}b).} So far every galaxy sat at the centre of its cutout. Shift it by $12$ zero-padded cardinal pixels and the ranking changes: Orbit-union loses $27.5$~pp while still never flipping under $D_4$, Flatten drops $25.8$~pp ($D_4$ flips $12.5\to 33.2\%$), and hard Radial rings fall $18.1$~pp. Is the culprit the hard ring edges or the fixed centre? Fixed Gaussian separates the two: smooth fixed-centre rings recover $+6.12$~pp over hard annuli at $12$ pixels ($78.33\%$), yet still fall $12.67$~pp from $\Delta=0$. Anchoring those same rings to the activation centroid holds mean accuracy at $89.68$--$90.07\%$ with zero $D_4$ flips. At $12$ pixels, Centred Gaussian beats Fixed Gaussian by $11.52$~pp (both trained end to end; conditional $95\%$ interval $9.30$--$13.74$~pp; App.~Table~\ref{tab:app_shifts}). The fix is not complete: jointly across all four directions, $87.27\%$ of galaxies stay correct, and $47/833$ change label in at least one direction relative to $\Delta=0$.

\section{Discussion, Limitations, and Conclusion}
\label{sec:conclusion}

Equivariant convolutions are only half of an invariant classifier; the readout is the other half, and a single padding pixel can undo the first. On this LeNet geometry, an equivariant trunk (odd pre-decimation parity or equivariant subsampling~\cite{xu2021group,rahman2025group}) paired with an exact-$D_4$ readout, whether fixed ($MP_g=M$) or centroid-anchored, is sufficient for exact $D_4$ invariance. With enough updates, compact invariant readouts approach flattened accuracy with over $100\times$ fewer head weights, but only rings that follow the activations survive off-centre cutouts. So turn the cutout to check whether the prediction stays fixed, then shift it to check whether the classifier still recognises the source. \textbf{Limitations:} we audit commit~\texttt{90d104b3}~\cite{scaife2021code}, not the private checkpoints behind Table~5 of Scaife and Porter~\cite{scaife2021fanaroff}; exactness covers $D_4$ only (other angles incur interpolation error); bright neighbours or clipped lobes can pull a flux centroid; out-of-fold tests are conditional on overlapping CV fits; and deeper \texttt{escnn} backbones remain future work.

\clearpage
\bibliographystyle{unsrtnat}
\bibliography{references}

\clearpage
\appendix

\section{Full Proofs and \texttt{escnn} Source Check}
\label{app:proofs}

\paragraph{Proof of Proposition~\ref{prop:readout}.}
Let $h_j(F) = \sum_{c=1}^C w_{j,c}^\top \operatorname{vec}(F_c) + b_j$ be unit $j$ of the first affine layer and $(g\cdot F)_c = \operatorname{vec}^{-1}(P_g\operatorname{vec}(F_c))$ for $g\in D_4$. Then
\[
h_j(g\cdot F) - h_j(F) = \sum_{c=1}^C w_{j,c}^\top (P_g - I)\operatorname{vec}(F_c) = \sum_{c=1}^C \bigl((P_g^\top - I)w_{j,c}\bigr)^\top \operatorname{vec}(F_c).
\]
Setting $F_{c'}=0$ for all $c'\neq c$ and varying $\operatorname{vec}(F_c)\in\mathbb{R}^{S^2}$ arbitrarily shows that this vanishes on all of $\mathbb{R}^{C\times S\times S}$ if and only if $(P_g^\top - I)w_{j,c}=0$ for every $j,c,g$, i.e., $w_{j,c}$ is constant on every $D_4$-orbit. Conversely, when a trunk satisfies $\operatorname{vec}(F_c(g\cdot x))=P_g\operatorname{vec}(F_c(x))$, any fixed pooling matrix $M\in\mathbb{R}^{K\times S^2}$ with $MP_g=M$ yields $M\operatorname{vec}(F_c(g\cdot x)) = M\operatorname{vec}(F_c(x))$ identically in real arithmetic, guaranteeing exact $D_4$ invariance for any subsequent evaluation-mode MLP. \hfill$\square$

\paragraph{Exact $D_4$ invariance of the Centred Gaussian readout.}
For any $D_4$-equivariant scalar-field map $F$ on $\{0,\dots,S-1\}^2$, $F(g\cdot x)(g\cdot u)=F(x)(u)$ for all $g\in D_4$ acting by orthogonal reflection/rotation about $c_0=((S-1)/2,(S-1)/2)$. The channel-averaged energy $\bar{F}(u)=\frac{1}{C}\sum_c \operatorname{ReLU}(F_c(u))+10^{-6}>0$ and normalised weights $w(u)=\bar{F}(u)/\sum_v \bar{F}(v)$ satisfy $w_{g\cdot x}(g\cdot u)=w_x(u)$. Because $g$ is an affine isometry on $\mathbb{R}^2$, $c(F(g\cdot x)) = \sum_{u'} w_{g\cdot x}(u')\,u' = \sum_u w_x(u)\,(g\cdot u) = g\cdot c(F(x))$. Hence the protected distance $r(u)=\sqrt{\|u-c(F)\|_2^2+10^{-8}}$ satisfies $r_{g\cdot x}(g\cdot u)=r_x(u)$, so the Gaussian ring weights $\psi_k(u)=\exp(-(r(u)-\mu_k)^2/(2\sigma^2))$ (normalised with a $10^{-8}$ floor) satisfy $\tilde{\psi}_{k,g\cdot x}(g\cdot u)=\tilde{\psi}_{k,x}(u)$, making every ring average strictly $D_4$-invariant in real arithmetic (verified on untrained $D_4$ trunks over $N=8$ cutouts: $\max_i \operatorname{std}_{D_4}(m_i) = 1.94\times 10^{-7}$ for Fixed Gaussian and $2.77\times 10^{-7}$ for Centred Gaussian, $0/8$ flips).

\paragraph{Stride-2 parity in \texttt{e2cnn v0.2.3} and \texttt{escnn v1.0.11}.}
Source inspection of \texttt{escnn v1.0.11} (\texttt{escnn/nn/modules/pooling/pointwise\_max.py}, \texttt{PointwiseMaxPoolAntialiased2D}) confirms the same two-stage operator as \texttt{e2cnn v0.2.3}: dense $2\times 2$ max pooling ($S-2\mapsto S-3$) followed by symmetric $5\times 5$ Gaussian blur ($\sigma=0.6$, pad~2) with stride~2 sampling even indices. For native $S=150$, $L=147$ is odd, so $i\mapsto 146-i$ preserves even phase ($150\to 148\to 147\to 74\to 72\to 71\to 36$). For padded $S=151$, $L=148$ is even, so $i\mapsto 147-i$ swaps even and odd indices at \texttt{Pool1}. The forward path is traced from \texttt{models.py} (lines 155--230, flatten at line 223) and \texttt{main.py} (lines 50--61 and 88). The released model also builds a circular input mask at initialisation but never calls it in \texttt{forward()}.

\clearpage
\section{Full Stage-Wise Parity Audit and Untrained Head Check}
\label{app:audit}

\begin{table}[!htbp]
\centering
\caption{Stage-wise exact-$D_4$ trunk audit and untrained head check. Top: mean (max over 7 non-identity $D_4$ actions at \texttt{GPool}) batch-norm relative error with shared conv weights ($N=16$; $^\dagger$equal-batch $N=4$ control). Bottom: MLP head parameters and untrained exact-$D_4$ logit std (flip rate) on $150^2$ and $151^2$ ($N=16$; $^\dagger N=4$ for Orbit-union; $^\ddagger$max over $N=8$ for Fixed and Centred Gaussian; `---' = not evaluated).}
\label{tab:app_audit}
\scriptsize
\setlength{\tabcolsep}{3.0pt}
\begin{tabular}{@{}llccccc@{}}
\toprule
\textbf{Arch} & \textbf{Grid / Stride} & \textbf{Conv1} & \textbf{Pool1} & \textbf{Conv2} & \textbf{Pool2} & \textbf{GPool (Max)} \\
\midrule
$D_4$ & $150^2$ ($s=2$) & $1.07\mathrm{e}{-7}$ & $1.19\mathrm{e}{-7}$ & $7.26\mathrm{e}{-7}$ & $4.60\mathrm{e}{-7}$ & $4.58\mathrm{e}{-7}$ ($5.37\mathrm{e}{-7}$) \\
$D_4$ & $151^2$ (pad, $s=2$) & $1.07\mathrm{e}{-7}$ & $2.49\mathrm{e}{-1}$ & $2.35\mathrm{e}{-1}$ & $1.68\mathrm{e}{-1}$ & $1.36\mathrm{e}{-1}$ ($1.96\mathrm{e}{-1}$) \\
$D_4^\dagger$ & $151^2$ ($s=1$ ctrl) & $1.06\mathrm{e}{-7}$ & $1.20\mathrm{e}{-7}$ & $7.82\mathrm{e}{-7}$ & $4.56\mathrm{e}{-7}$ & $4.53\mathrm{e}{-7}$ ($5.14\mathrm{e}{-7}$) \\
$D_{16}$ & $150^2$ ($s=2$) & $1.12\mathrm{e}{-7}$ & $1.20\mathrm{e}{-7}$ & $1.39\mathrm{e}{-6}$ & $8.66\mathrm{e}{-7}$ & $8.68\mathrm{e}{-7}$ ($1.03\mathrm{e}{-6}$) \\
$D_{16}$ & $151^2$ (pad, $s=2$) & $1.12\mathrm{e}{-7}$ & $2.57\mathrm{e}{-1}$ & $2.23\mathrm{e}{-1}$ & $1.62\mathrm{e}{-1}$ & $1.28\mathrm{e}{-1}$ ($1.85\mathrm{e}{-1}$) \\
$D_{16}^\dagger$ & $151^2$ ($s=1$ ctrl) & $1.12\mathrm{e}{-7}$ & $1.21\mathrm{e}{-7}$ & $1.48\mathrm{e}{-6}$ & $8.39\mathrm{e}{-7}$ & $8.54\mathrm{e}{-7}$ ($9.64\mathrm{e}{-7}$) \\
\bottomrule
\end{tabular}
\vspace{5pt}

\begin{tabular}{@{}lrrcccc@{}}
\toprule
\textbf{Head} & \textbf{Dim} & \textbf{Params} & \textbf{$D_4$, $150^2$} & \textbf{$D_4$, $151^2$} & \textbf{$D_{16}$, $150^2$} & \textbf{$D_{16}$, $151^2$} \\
\midrule
Flatten & $20{,}736$ & $2{,}498{,}774$ & $0.340\ (12.5\%)$ & $0.333\ (25.0\%)$ & $0.176\ (62.5\%)$ & $0.168\ (43.8\%)$ \\
GAP & $16$ & $12{,}374$ & $1.95\mathrm{e}{-8}\ (0\%)$ & $6.25\mathrm{e}{-4}\ (0\%)$ & $1.86\mathrm{e}{-8}\ (0\%)$ & $2.00\mathrm{e}{-4}\ (0\%)$ \\
Global statistics & $96$ & $21{,}974$ & $1.05\mathrm{e}{-6}\ (0\%)$ & $5.55\mathrm{e}{-2}\ (0\%)$ & $9.91\mathrm{e}{-7}\ (0\%)$ & $2.22\mathrm{e}{-2}\ (0\%)$ \\
Orbit-union$^\dagger$ & $96$ & $21{,}974$ & $6.29\mathrm{e}{-8}\ (0\%)$ & $5.36\mathrm{e}{-3}\ (0\%)$ & $4.07\mathrm{e}{-8}\ (0\%)$ & $3.48\mathrm{e}{-3}\ (0\%)$ \\
Radial ($K=6$) & $96$ & $21{,}974$ & $2.78\mathrm{e}{-7}\ (0\%)$ & $9.54\mathrm{e}{-3}\ (0\%)$ & $1.92\mathrm{e}{-7}\ (0\%)$ & $1.75\mathrm{e}{-2}\ (0\%)$ \\
Fixed Gaussian$^\ddagger$ ($K=6$) & $96$ & $21{,}974$ & $1.94\mathrm{e}{-7}\ (0\%)$ & --- & --- & --- \\
Centred Gaussian$^\ddagger$ ($K=6$) & $96$ & $21{,}974$ & $2.77\mathrm{e}{-7}\ (0\%)$ & --- & --- & --- \\
\bottomrule
\end{tabular}
\end{table}

\FloatBarrier
\section{Complete Paired Statistical Contrasts and $2\times 5$ Crossed Frozen-Trunk Table}
\label{app:stats}

\begin{table}[!htbp]
\centering
\caption{Long-budget end-to-end paired contrasts ($N=833$, $120$ epochs, batch size $64$, $1{,}080$ steps). Mean best validation epochs across the five folds: Flatten $36.2$, Radial $59.4$, Orbit-union $72.4$, Fixed Gaussian $80.4$, Centred Gaussian $91.8$, GAP $95.8$. Unadjusted conditional paired bootstrap $95\%$ CIs, exact two-sided McNemar $p$-values (Holm-adjusted over the 10-contrast family), discordant wins/losses, and per-fold net wins.}
\label{tab:app_w1_contrasts}
\scriptsize
\setlength{\tabcolsep}{3.2pt}
\begin{tabular}{@{}lcccc@{}}
\toprule
\textbf{Contrast (Long-Budget E2E)} & \textbf{$\Delta$Acc [95\% CI] (pp)} & \textbf{$p$ (Holm)} & \textbf{W / L} & \textbf{Fold Net Wins} \\
\midrule
GAP $-$ Flatten & $-5.16\ [-7.80, -2.52]$ & $1.50\mathrm{e}{-4}\ (1.35\mathrm{e}{-3})$ & $41 / 84$ & $[-7, -8, -8, -10, -10]$ \\
Orbit-union $-$ GAP & $+4.80\ [+2.52, +7.08]$ & $5.46\mathrm{e}{-5}\ (5.46\mathrm{e}{-4})$ & $68 / 28$ & $[+6, +5, +14, +11, +4]$ \\
Fixed Gaussian $-$ GAP & $+3.96\ [+1.68, +6.36]$ & $1.33\mathrm{e}{-3}\ (0.0107)$ & $67 / 34$ & $[+4, +10, +5, +8, +6]$ \\
Radial $-$ GAP & $+3.24\ [+0.84, +5.64]$ & $0.0124\ (0.0868)$ & $68 / 41$ & $[+3, +6, +9, +7, +2]$ \\
Centred Gaussian $-$ GAP & $+2.64\ [+0.36, +4.92]$ & $0.0280\ (0.1681)$ & $57 / 35$ & $[+6, +2, +3, +10, +1]$ \\
Orbit-union $-$ Flatten & $-0.36\ [-2.52, +1.80]$ & $0.8243\ (0.8243)$ & $39 / 42$ & $[-1, -3, +6, +1, -6]$ \\
Fixed Gaussian $-$ Flatten & $-1.20\ [-3.12, +0.72]$ & $0.2750\ (0.5500)$ & $29 / 39$ & $[-3, +2, -3, -2, -4]$ \\
Radial $-$ Flatten & $-1.92\ [-3.72, -0.12]$ & $0.0479\ (0.1918)$ & $21 / 37$ & $[-4, -2, +1, -3, -8]$ \\
Centred Gaussian $-$ Flatten & $-2.52\ [-4.80, -0.24]$ & $0.0375\ (0.1876)$ & $36 / 57$ & $[-1, -6, -5, 0, -9]$ \\
Centred Gaussian $-$ Fixed Gaussian & $-1.32\ [-3.00, +0.36]$ & $0.1690\ (0.5069)$ & $21 / 32$ & $[+2, -8, -2, +2, -5]$ \\
\bottomrule
\end{tabular}
\end{table}

\begin{table}[!htbp]
\centering
\caption{Complete $2\times 5$ crossed frozen-trunk ablation ($N=833$, exact unclipped logit NLL). All five heads refitted on both frozen Flatten- and Radial-trained trunks under matched short protocol ($300$ steps, $O(2)$) vs.\ 8-view schedule ($3.4\text{k}$ steps). $D_4$/PIL: exact-$D_4$ and 32-view PIL flip counts out of $833$; $n$: galaxies correct. Bold: best exact-$D_4$ head per column within each trunk; the all-zero $D_4$ column is unmarked.}
\label{tab:app_crossed}
\scriptsize
\setlength{\tabcolsep}{3.4pt}
\begin{tabular}{@{}llccccccc@{}}
\toprule
& & \multicolumn{3}{c}{\textbf{Matched Protocol ($300$ steps)}} & \multicolumn{4}{c}{\textbf{8-View Schedule ($3.4\text{k}$ steps)}} \\
\cmidrule(lr){3-5}\cmidrule(l){6-9}
\textbf{Trunk} & \textbf{Head} & \textbf{Acc (\%)} & \textbf{NLL} & \textbf{$D_4$/PIL} & \textbf{Acc ($n$)} & \textbf{NLL} & \textbf{$D_4$} & \textbf{PIL32} \\
\midrule
\multirow{5}{*}{Flatten}
  & Flatten & $91.36$ & $0.217$ & $94/144$ & $91.00\ (758)$ & $0.351$ & $91$ & $179$ \\
  & GAP & $83.91$ & $0.474$ & $0/108$ & $84.27\ (702)$ & $0.366$ & $0$ & $97$ \\
  & Global statistics & $81.99$ & $0.417$ & $0/151$ & $85.35\ (711)$ & $0.342$ & $0$ & $119$ \\
  & Orbit-union & $87.27$ & $0.315$ & $0/93$ & $89.08\ (742)$ & $0.296$ & $0$ & $115$ \\
  & Radial ($K=6$) & $\mathbf{88.00}$ & $\mathbf{0.286}$ & $\mathbf{0/59}$ & $\mathbf{91.60}\ (763)$ & $\mathbf{0.251}$ & $0$ & $\mathbf{57}$ \\
\midrule
\multirow{5}{*}{Radial}
  & Flatten & $89.56$ & $0.270$ & $127/171$ & $88.84\ (740)$ & $0.380$ & $106$ & $202$ \\
  & GAP & $80.19$ & $0.484$ & $0/129$ & $84.03\ (700)$ & $0.370$ & $0$ & $87$ \\
  & Global statistics & $79.59$ & $0.452$ & $0/136$ & $86.07\ (717)$ & $0.351$ & $0$ & $136$ \\
  & Orbit-union & $86.19$ & $0.341$ & $0/91$ & $\mathbf{90.04}\ (750)$ & $\mathbf{0.275}$ & $0$ & $97$ \\
  & Radial ($K=6$) & $\mathbf{88.24}$ & $\mathbf{0.299}$ & $\mathbf{0/45}$ & $89.80\ (748)$ & $0.343$ & $0$ & $\mathbf{48}$ \\
\bottomrule
\end{tabular}
\end{table}

\FloatBarrier
\section{Off-Centre Shift Stress Test Details and Reproducibility}
\label{app:shifts}

\begin{table}[!htbp]
\centering
\caption{Full off-centre cutout shift breakdown ($\Delta\in\{0,2,4,8,12\}$~px; $833$ canonical at $\Delta=0$, $3{,}332$ shifted evals per nonzero $\Delta$). Accuracy (\%) and exact-$D_4$ flip rate (\%) at each shift, plus conditional per-galaxy paired $\text{mean}\pm 1.96\,\text{SE}$ intervals for Centred Gaussian minus each baseline at $\Delta=12$~px. Bold: best exact-$D_4$ head per shift.}
\label{tab:app_shifts}
\scriptsize
\setlength{\tabcolsep}{1.5pt}
\begin{tabular}{@{}lcccccc@{}}
\toprule
\textbf{Model} & $\Delta{=}0$~px & $\Delta{=}2$~px & $\Delta{=}4$~px & $\Delta{=}8$~px & $\Delta{=}12$~px & \textbf{Centred $-$ Model ($\Delta{=}12$)} \\
\midrule
$D_4$ Flatten & $92.20\ (12.5\%)$ & $90.37\ (13.6\%)$ & $83.28\ (18.3\%)$ & $66.39\ (22.6\%)$ & $66.45\ (33.2\%)$ & $+23.41 \pm 2.48$~pp \\
$D_4$ GAP & $87.03\ (0.0\%)$ & $86.52\ (0.0\%)$ & $87.15\ (0.0\%)$ & $87.15\ (0.0\%)$ & $86.88\ (0.0\%)$ & $+2.97 \pm 2.01$~pp \\
$D_4$ Orbit-union & $\mathbf{91.84}\ (0.0\%)$ & $\mathbf{90.43}\ (0.0\%)$ & $84.66\ (0.0\%)$ & $67.35\ (0.0\%)$ & $64.32\ (0.0\%)$ & $+25.54 \pm 2.85$~pp \\
$D_4$ Radial & $90.28\ (0.0\%)$ & $90.19\ (0.0\%)$ & $89.56\ (0.0\%)$ & $78.66\ (0.0\%)$ & $72.21\ (0.0\%)$ & $+17.65 \pm 2.27$~pp \\
$D_4$ Fixed Gaussian & $91.00\ (0.0\%)$ & $90.28\ (0.0\%)$ & $89.74\ (0.0\%)$ & $86.37\ (0.0\%)$ & $78.33\ (0.0\%)$ & $+11.52 \pm 2.22$~pp \\
$D_4$ Centred Gaussian & $89.68\ (0.0\%)$ & $89.83\ (0.0\%)$ & $\mathbf{89.92}\ (0.0\%)$ & $\mathbf{90.07}\ (0.0\%)$ & $\mathbf{89.86}\ (0.0\%)$ & --- \\
\bottomrule
\end{tabular}
\end{table}

\paragraph{Training and reproducibility details.} Experiments used \texttt{PyTorch 2.15.0}, \texttt{NumPy 1.26.4}, and \texttt{e2cnn 0.2.3} on Apple Silicon MPS with fixed seeds. MiraBest contains $1{,}222$ unique MD5 hashes normalised with $\mu=0.0031, \sigma=0.0350$~\cite{scaife2021code} ($0.00$ PIL-$0^\circ$ identity error). End-to-end folds use an $18\%$ validation split, Adam ($\text{lr}=10^{-3}$, cosine decay, weight decay $10^{-5}$), and continuous $O(2)$ augmentation on $150^2$ (batch size $110$ for $60$-epoch runs; $64$ for $120$-epoch runs). Saved CSVs in \texttt{results/} store per-galaxy canonical logits, all eight exact $D_4$ margins, and 32 PIL margins for long-budget, Fixed Gaussian, and multi-seed runs; \texttt{audit\_workshop\_tables.py} in \texttt{experiments/} recomputes the main accuracy, flip rates, and paired contrasts without retraining.

\end{document}